%% file: review_generation_aaai17.tex
\def\year{2017}\relax
\documentclass[letterpaper]{article}
\usepackage{aaai17}
\usepackage{times}
\usepackage{helvet}
\usepackage{courier}
\usepackage{times}
\usepackage{latexsym}
\usepackage{subfigure,graphicx,eqnarray,amsmath,multirow,multicol,url}

\newtheorem{definition}{Definition}

\begin{document}
%
\title{Context-aware Natural Language Generation with Recurrent Neural Networks}
\author{Jian Tang$^1$, Yifan Yang$^2$, Sam Carton$^3$, Ming Zhang$^2$, and Qiaozhu Mei$^3$\\
$^1$ Microsoft Research, $^2$ Peking University, $^3$ University of Michigan\\
\{tangjianpku,yang1fan2\}@gmail.com, mzhang\_cs@pku.edu.cn, \{scarton, qmei\}@umich.edu
}
\maketitle

\begin{abstract}
	This paper studied generating natural languages at particular contexts or situations. We proposed two novel approaches which encode the contexts into a continuous semantic representation and then decode the semantic representation into text sequences with recurrent neural networks. During decoding, the context information are attended through a gating mechanism, addressing the problem of long-range dependency caused by lengthy sequences. We evaluate the effectiveness of the proposed approaches on user review data, in which rich contexts are available and two informative contexts, sentiments and products, are selected for evaluation. Experiments show that the fake reviews generated by our approaches are very natural. Results of fake review detection with human judges show that more than 50\% of the fake reviews are misclassified as the real reviews, and more than 90\% are misclassified by existing state-of-the-art fake review detection algorithm.
\end{abstract}

\input{introduction}
\input{related}
\input{definition}
\input{model}

\input{experiment}

\input{conclusion}

\bibliographystyle{aaai}
\bibliography{sigproc}

\end{document}

%% file: introduction.tex
\section{Introduction}
\label{sec::intro}
\begin{figure*}[htbp]
	\centering
	\label{fig::intro}
	\includegraphics[width=0.7\textwidth]{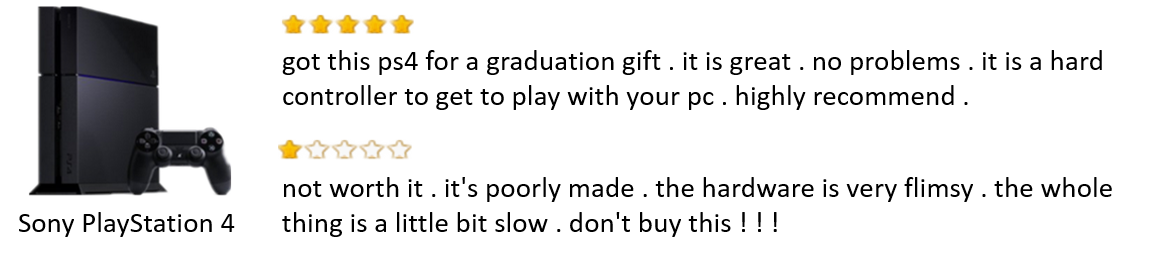}
	\caption{Examples of reviews generated by our approach. Only the \emph{sentiment rating} and \emph{product id} are fed to the algorithm. }
\end{figure*}
Natural language generation is potentially useful in a variety of applications such as natural language understanding~\cite{graves2013generating}, response generation in dialogue systems~\cite{wen2015stochastic,wen2015semantically,sordoni2015neural}, text summarization~\cite{rush2015neural}, machine translation~\cite{bahdanau2014neural} and image caption~\cite{xu2015show}. Traditional approaches usually generate languages according to some \emph{rules} or \emph{templates} designed by humans~\cite{cheyer2014method}, which are specific for some tasks and domains and difficult to generalize to other tasks and domains. Besides, the languages generated according to these approaches are very similar, lacking the large variations of human languages. Therefore, it is a long shot of the community to develop automatic approaches that learn from data and generate languages as diverse as human languages. 

Recently, recurrent neural networks (RNNs) have been proved to very effective in natural language generation~\cite{graves2013generating,sutskever2011generating,bowman2015generating}. Comparing to the traditional approaches, RNNs directly model the generation process of text sequences, and the generating function can be automatically learned from a large amount of text data, providing an end-to-end solution. Though traditional RNNs suffer from the problem of gradient vanishing or exploding, the long-short term memory (LSTM)~\cite{hochreiter1997long} unit effectively addresses this problem and is able to capture the long-range dependency in natural languages. RNNs with LSTM have shown very promising results on various data sets with different structures including Wikipedia articles~\cite{graves2013generating}, linux source codes~\cite{karpathy2015visualizing}, scientific papers~\cite{karpathy2015visualizing}, NSF abstracts~\cite{karpathy2015visualizing}. 

Most of existing work focus on natural language generation only with their textual content while ignoring their contextual information.  However, in reality natural languages are usually generated at/with particular contexts, e.g., time, locations, emotions or sentiments. Therefore, in this paper we study \emph{context-aware natural language generation}. Our goal is to generate not only semantically and syntactically coherent sentences, but also sentences that are reasonable at particular contexts. Indeed, contexts have been proved to be very useful for various natural language processing tasks such as topic extraction~\cite{mei2006probabilistic}, text classification~\cite{cao2009context} and language modeling{mikolov2012context}.

We proposed two novel approaches for context-aware natural language generation, which map a set of contexts to text sequences. Our first model C2S encodes a set of contexts into a continuous representation and then decode the representation into a text sequence through a recurrent neural network. The C2S model is able to generate semantically and syntactically very coherent sentences. However, one limitation is that when the sequences become very long, the information from the contexts may not be able to propagate to the words in distant positions. An intuitive approach to address this is to build the direct dependency between the contexts and the words in the sequences, allowing the information jump from the contexts to the words. However, not all the words may depend on the contexts, some of which may only depend on their preceding words. To resolve this, a gating mechanism is introduced to control when the information from the contexts are accessed. This is our second model: gated contexts to sequences (gC2S). 

We evaluate our approaches on the user reviews from Amazon and TripAdvisor, where rich contexts are available. Two informative contexts are selected: sentiment rating and product id. Fig.~\ref{fig::intro} presents two examples of reviews generated by gC2S, which are very difficult to tell from reviews written by real users. We choose the task of fake review detection to evaluate the effectiveness of our approach. Experimental results show that more than 50\% of the fake reviews generated by our approach are misclassified as real reviews with human judges and more than 90\% are misclassified by the existing state-of-the-art fake review detection algorithm. 

%% file: related.tex
\section{Related Work}
\label{sec::related}
The approaches of natural language generation can be roughly classified into two categories: the classical rule-based or template-based approaches, and recent approaches with recurrent neural networks, which automatically learn the natural language generator from the data.  Classical approaches usually define some rules or templates~\cite{cheyer2014method} by humans, which are very brittle and hard to generalize to different tasks and domains. Though there are some recent approaches aiming to learn the template structures from large amounts of corpus~\cite{oh2000stochastic}, the training data is very expensive to obtain and the final generation process still requires additional human handcrafted features. 

Comparing to the traditional rule-based approaches, the recurrent neural networks based approaches does not rely on any human handcrafted features and provide an end-to-end solution. Our approach is also built on recurrent neural networks. \cite{graves2013generating} studied sequence generation, including text, using recurrent neural networks (RNN) with long-short term memory unit. \cite{sutskever2011generating} proposed a multiplicative RNN (MRNN) for text prediction and generation, in which different transformation functions between the hidden states are used for different input characters. \cite{bowman2015generating} investigated generating sentences from continuous semantic spaces with a variational auto-encoder, in which RNN is used for both the encoder and the encoder. These work have shown that RNNs are very effective for text generation on various data sets of different structures. However, all these work study natural language generation without contexts.

There are some recent work that investigate language modeling with context information. \cite{mikolov2012context} studied language modeling by adding the topical features of preceding words as contexts. \cite{wang2015larger} exploited the preceding words in larger windows for language modeling. These work focus on the task of language modeling and the preceding words are used as the contexts information while our task focuses on natural language generation and external contexts are studied.
There are also some related work of response generation in conversation systems~\cite{sordoni2015neural,wen2015semantically}, in which the conversation history are treated as contexts.  Comparing to their work, our solutions are more general, application for a variety of context while their solution are specifically designed for contexts with specific structures.

%% file: definition.tex
\section{Problem Definition}
\label{sec::definition}

Natural languages are usually associated with rich context information, e.g., time, location, which provide clues on how the natural languages are generated. In this paper, we study \emph{context-aware} natural language generation. Given the context clues, we want to generate the corresponding natural languages. We first formally define the contexts as follows:

\begin{definition}
	\label{def::contexts}
	\textbf{(Contexts.) }
	\textsl{The \textbf{contexts} of natural languages refer to the situations they are generated. Each context is defined as a high-dimensional vector.}
\end{definition}

The contexts of natural languages can be either discrete or continuous features. For example, the context can be a specific user or location; it can also be a continuous feature vectors generated from other sources. For discrete features, the context is usually represented with one-hot representations. Formally, we formulate our problem as follows:


\begin{definition}
	\label{def::problem set}
	\textbf{(Context-aware natural language generation)}
	\textsl{ Given a set of contexts $C=\{\vec{c}_i\}_{i=1,\ldots, K}$, in which $K$ is the total number of context types, our goal is to generate a sequence of words $w_1,w_2,\ldots,w_n$ that are appropriate at the given contexts }
\end{definition}

In this paper, we take the user reviews as an example, where there exist abundant context information, e.g., user, time, sentiment, product. However, our proposed approaches are also general for other contextual data. Next, we introduce our approach for context-aware natural language generation.

%% file: model.tex
\section{Model}
\label{sec::model}
In this section, we introduce our proposed approaches for generating natural language at particular contexts. We first introduce the recurrent neural networks (RNN), which are very effective models for text generation, and then introduce our proposed approaches, which map a set of contexts to a text sequence. 

\begin{figure*}[htbp]
	\centering
	\subfigure[RNN for sequence modeling]{
		\label{fig::rnn}
		\includegraphics[width=0.3\textwidth]{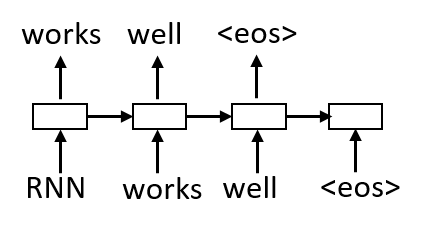}
	}
	\subfigure[C2S: Contexts to Sequences]{
		\label{fig::c2s}
		\includegraphics[width=0.3\textwidth]{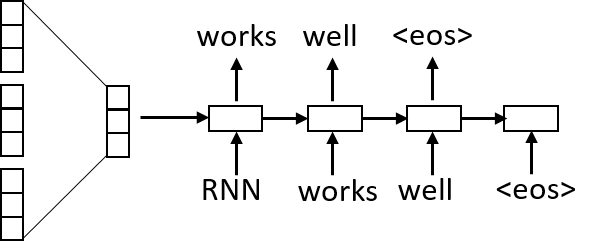}
	}
	\subfigure[gC2S: Gated Contexts to Sequences]{
		\label{fig::gc2s}
		\includegraphics[width=0.3\textwidth]{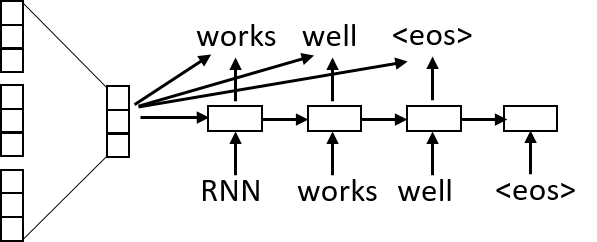}
	}			
	\caption{(a): classical RNN model for text modeling; (b): our first approach in which multi-layer neural networks are encoders and RNN are decoders; (c): our second approach which adds skip-connections from contexts to words, controlled by a gating function. }
	\label{fig::models}
\end{figure*}

\subsection{Recurrent Neural Network}
Recurrent neural network (RNN) models the generative process of sequence data, which summarizes the information into a hidden state ( a continuous representation) and then generate a new sample according to a probability distribution specified by the hidden state. Specifically, the hidden state $h_t$ from the sequence $x_1,x_2,\ldots, x_t$ is recursively updated as:
\begin{equation}
h_t=f(h_{t-1}, x_t),
\end{equation}
where $f(\cdot,\cdot)$ is usually a nonlinear transformation, e.g., $h_t=\tanh(Uh_{t-1}+Vx_t))$ ($U,V$ are transformation matrices). The hidden state $h_t$ summarizes the information of the entire sequences $x_1,x_2,\ldots, x_t$, and the probability of generating next words $p(x_{t+1}|x_{\leq t})$ is defined as
\begin{equation}
p(x_{t+1}|x_{\leq t})=p(x_{t+1}|h_t) \propto \exp(O_{x_{t+1}}^T h_t),
\end{equation}
where $O_{x_{t+1}}$ is the low-dimensional continuous representation of word $x_{t+1}$. 

The overall probability of a sequence $\vec{x}={x_1,x_2,\ldots, x_T}$ is calculated as follows: 
\begin{equation}
\label{eqn::joint_prob}
p(\vec{x})=\prod_{t=1}^T p(x_t|h_{t-1}),
\end{equation}

Training RNN can be done through maximizing the joint probability $p(x)$ defined by Eqn.~\eqref{eqn::joint_prob} and optimized through Back-propagation. However, training RNN with traditional state transition unit $h_t=\tanh(Ux_t+Vh_{t-1})$ suffers from the problem of gradient vanishing or exploding, which is caused by the product of multiple non-linear transforming functions. \cite{hochreiter1997long} effectively addresses this problem through the long-short term memory (LSTM) unit. The core idea of LSTM is introducing the memory state and multiple gating functions to control the information written to the memory sate, reading from the memory state, and removed (or forgotten) from the memory state. Specifically, the detailed updating equations are listed below:
\begin{equation}
\begin{aligned}
	z_t&=\tanh(W_zx_t+W_hh_{t-1}+b_z)\\
	i_t&=\sigma(W_ix_t+W_hh_{t-1}+b_i) \\
	f_t&=\sigma(W_fx_t+W_hh_{t-1}+b_f) \\
	c_t&=f_t\odot c_{t-1}+i_t\odot z_t \\
	o_t&=\sigma(W_ox_t+W_hh_{t-1}+b_o) \\
	h_t&=o_t\odot\tanh(c_t),		
\end{aligned}	
\end{equation}
where $c_t$ is the memory state, $z_t$ is the module that transform information from input space $x_t$ to the memory space, $h_t$ is the information read from the memory state, $i_t, f_t, o_t$ are the input, forget, and output gates respectively. $i_t$ controls the information from input $z_t$ to the memory state, $f_t$ controls the information in the memory state to be forgotten, $o_t$ controls the information read from the memory state. The memory state $c_t$ is updated through a linear combination of the input filtered by the input gate and the previous memory state filtered by the forget gate.

\subsection{C2S: Contexts to Sequences}
RNN effectively the joint probability of natural languages $p(\vec{x})$. As mentioned previously, natural languages are usually generated at particular contexts.Therefore, instead of modeling the probability of observing a text sequence $p(\vec{x})$, we are more interested in the probability of observing $x$ under some contexts $C$, i.e., the conditional probability $p(\vec{x}|C)$. In this part, we introduce two generative models for modeling the conditional probability $p(\vec{x}|C)$ based on recurrent neural networks. 

\noindent\textbf{Encoder.} Our framework is built on the encoder-decoder framework~\cite{cho2014learning}. The essential idea is to encode the contexts information into a continuous semantic representation, and then decode the semantic representation into a text sequence. We first introduce how to encode the contexts of natural languages into a semantic representation. We represent the contexts as a set $C=\{\vec{c}_i\}_{i=1, K}$, where $\vec{c}_i$ is a type of context, $K$ is the number of context types. Take the review as an example, each $\vec{c}_i$ is a sentiment rating score (ranging from 1 to 5), a product id or a user id. For discrete contexts, each $\vec{c}_i$ is a one hot-vector $\vec{c}_i$. The embedding of each context $\vec{c}_i$ can be obtained through:
\begin{equation}
\label{eqn::context_embedding}
 \vec{\vec{e}}_i=E_i\vec{\vec{c}}_i, 
\end{equation}
where $E_i \in R^{d\times|K_i|}$, $K_i$ is the number of different context values of type $i$, and $d$ is the dimension of context embedding. Once the embeddings of different contexts are obtained, they are concatenated into a long vector and followed by a non-linear transformation, formulated as follows:
\begin{equation}
\label{eqn::context_embedding_transformation}
	h_C=\tanh (W[\vec{e}_1, \vec{e}_2, \ldots, \vec{e}_{|C|}])+b),
\end{equation}
where $W\in R^{Kd\times N}$, $N$ is the size of hidden state of recurrent neural networks in the decoder. By Eqn.~\eqref{eqn::context_embedding} and Eqn.~\ref{eqn::context_embedding_transformation}, we are able to encoder the contexts into a semantic representation. Next we introduce how to decode it a text sequence. 

\noindent \textbf{Decoder.} We introduce two types of decoders. The first one is the vanilla recurrent neural networks with LSTM unit, and the initiate state of the RNN is set as the context embedding $h_C$. We call this approach as C2S, and the whole encoder-decoder framework is presented in~\ref{fig::c2s}. The C2S have shown very promising results in the experiments. However, one limitation of the approach is that when the sequences become very long, the information from the contexts may not be able to propagate to the distant words. To resolve this, a natural solution would be to directly build the dependency between the contexts $h_C$ and each word, i.e., add the skip-connections between the contexts and the words. By doing this, when predicting the next word $x_{t+1}$, it not only depends on the current hidden state $h_t$, but also depends on the context representation $h_C$. To combine the two sources of information, a simple way would be to take their summation or concatenate them. Here we use the way of taking their summation. However, simply summing the two representations which treats the two sources of information equally may be problematic as some words may depend on the context or others may not. Therefore, it would be a desirable to figure out an approach when the context information are required.

We achieve this through the gating mechanism. We introduce a gating function which depends on the current hidden state $h_t$:
\begin{equation}
	m_t=\sigma(Vh_t+b),
\end{equation} 
where $V\in R^{N\times N}$, $\sigma(\cdot)$ is the sigmoid function. The probability of next word $p(x_{t+1}|x_{\leq t}, C)$ will be calculated as follows:  
\begin{equation}
	p(x_{t+1}|x_{\leq t}, C)\propto \exp(O_{x_{t+1}}^T(h_t+m_t\odot h_C)),
\end{equation}
where $\odot$ is the elementwise product. We call this model gC2S, and the whole framework is presented in~\ref{fig::gc2s}. 

\noindent \textbf{Generation.} Once the models are trained, give a set of contexts, we can generate natural languages based on them. There are usually two types of approaches for natural language generation: beam search~\cite{bahdanau2014neural}, which is widely used in neural machine translation, and random sample~\cite{graves2013generating}. In our experiments, we tried both approaches. We find that the samples generated by the beam search are usually very trivial without much variation. Therefore, we adopt the approach of random sampling. During the sampling process, instead of using the standard softmax function, we also tried different values of temperatures. High temperatures will generate more diverse samples but making more mistakes while small temperatures tend to generate more conservative and confident samples. In the experiments, we empirically set the temperatures as 0.7.

%% file: experiment.tex
\section{Experiments}
\label{sec::experiment}
In this section, we evaluate the effectiveness of the approach C2S and gC2S with the user review data. Different tasks are evaluated including language modeling, fake review detection with human judges or existing state-of-the-art fake review detection algorithm, sentiment classification on both the real and fake reviews. We first introduce the data sets to be used. 

\begin{table}[!htdb]
	\caption{Statistics of the data sets}
	\label{tab::statistics}
	\centering
	\scalebox{0.7}{
		\begin{tabular}{c|c|c|c|c|c|c} \hline \hline
			Name&train& valid &test & Median.Len & Max Len & Vocab. \\ \hline
			Book& 5,800,000  & 293,050 & 293,055  &  22 &100  &20,000\\ \hline
			Electronic&3,200,000 & 180,708 & 180,709 & 35 &100  &20,000\\ \hline
			Movie & 1,500,000  & 85,195 & 85,200 & 31 & 100 &20,000 \\ \hline
			Hotel&230,000  & 12,912 & 12,914 & 87 & 300&20,000\\ \hline \hline
		\end{tabular}
	}
\end{table}

\begin{table*}[!htdb!]
	\caption{Results of language modeling measured by perplexity (P: product, S: sentiment).}
	\label{tab::results-lm}
	\centering
	\scalebox{0.9}{
		\begin{tabular}{c|c|c|c|c|c|c|c} \hline \hline
			Domain&RNN&C2S(P) & C2S(S)& C2S(P+S)&gC2S(P)&gC2S(S)&gC2S(P+S)  \\ \hline
			Book&27.5& 27.1  & 27.2 & 26.6 &25.2&25.8&\textbf{24.9} \\ \hline
			Electronic&27.4  &26.2  &27.3 &25.8 &24.4 &25.6&\textbf{24.1} \\ \hline
			Movie &28.8   &27.2  &28.2&26.9&25.3&27.1&\textbf{24.8}    \\ \hline
			Hotel&23.6  &23.2 &23.4 &23.1&21.3&22.4&\textbf{21.2}  \\ \hline \hline
		\end{tabular}
	}
\end{table*}

\begin{figure*}[!htbp]
	\centering
	\includegraphics[width=0.89\textwidth]{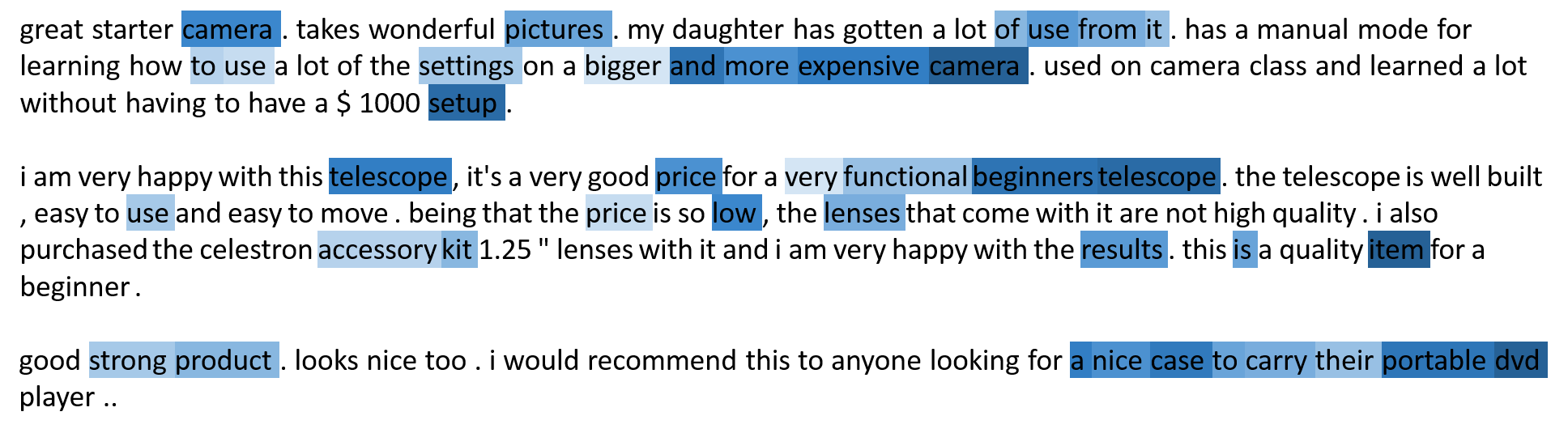}		
	\caption{Words with the largest gating values in the gC2S model on test data. Most of the words are product or sentiment related.  }
	\label{fig::gating-cs}
\end{figure*}

\subsection{Data Sets}
We choose the user review data as it contains rich context information, e.g., users, time, sentiments, products. We select the \textbf{sentiment ratings} (ranging from 1 to 5) and \textbf{product ids} as the context information, which we believe are the most important factors that affect the review content. We use data from two websites: \emph{Amazon}\footnote{Available at \url{http://jmcauley.ucsd.edu/data/amazon/links.html}} and \emph{TripAdvisor}\footnote{Available at \url{http://www.cs.cmu.edu/~jiweil/html/hotel-review.html}} are used. For the \emph{Amazon} data, we select three most popular domains including \emph{book}, \emph{electronic} and \emph{movie}; for the \emph{TripAdvisor} data, it is about the \emph{hotel} domain. We select the most popular $20,000$ words as the vocabulary, and reviews containing unknown words, with length more than 100 words in the \emph{Amazon} data and more than 300 words in the \emph{TripAdvisor} data are all removed. The whole data are split into train, validation, test data according to the ratio 18:1:1. The statistics of the final data sets are summarized into Table~\ref{tab::statistics}.

\noindent \textbf{Training Details.} All the models on trained on a single GPU. The batch size is set as 128. The weights are randomly initialized with the uniform distribution $(-0.1,0.1)$, and the biases are initialized with 0. The initial learning rate is set as 1, and the learning rate is halved if the perplexity of the current epoch on the validation data is not less than the last epoch. The gradient is clipped if the norm is larger than 5. Dropout is used from the input to hidden layer and from the hidden layer to output layer in the recurrent neural networks. Different values of hidden size is tried, and the results show that the larger, the better. Due to the limitation of GPU memory, we use 512 by default. For the number of layers of RNN, one layer is used by default as increasing the number of layers does not yield significantly better samples.
  
\begin{figure}[htbp]
	\centering
		\includegraphics[width=0.3\textwidth]{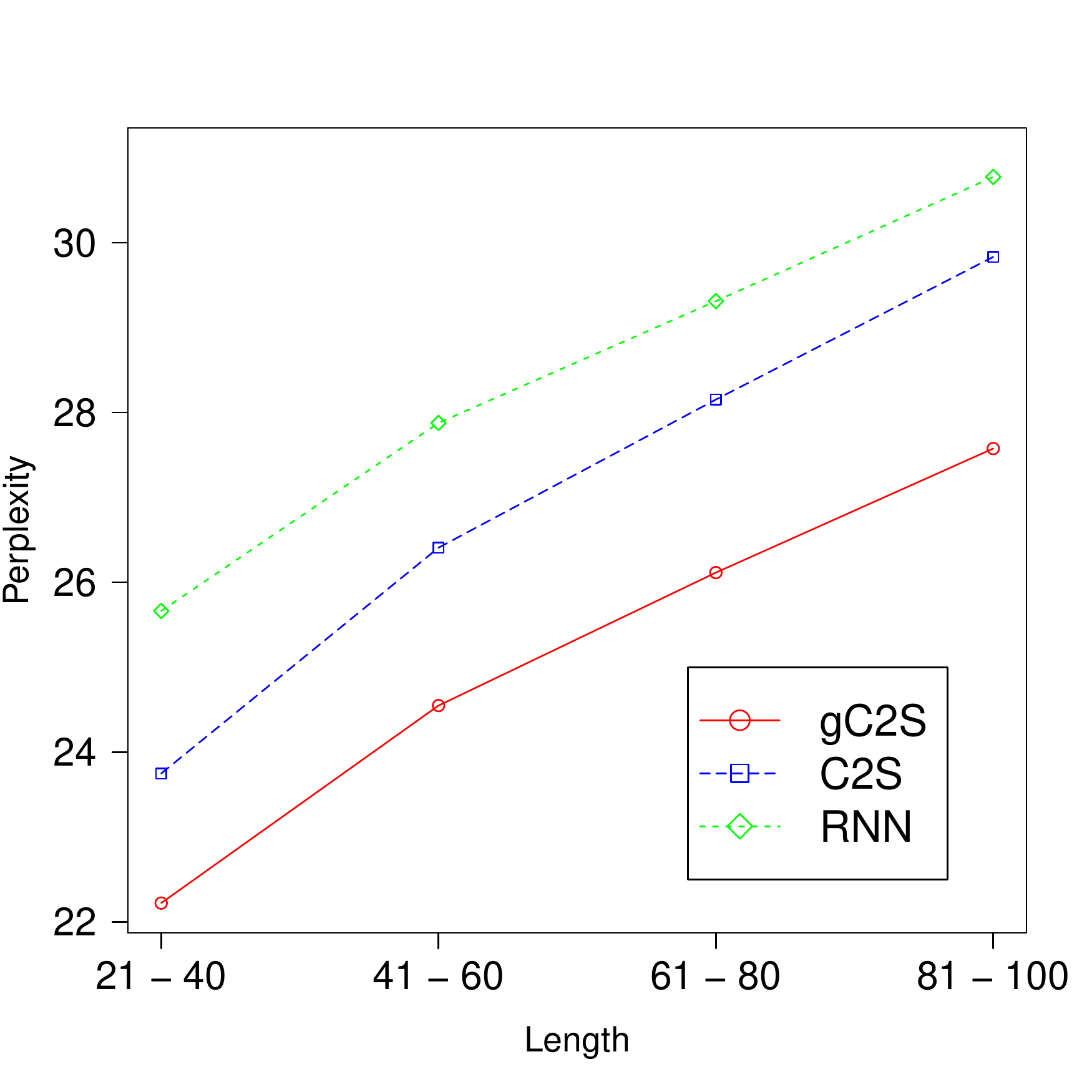}	
	\caption{Results of language modeling on different lengths of reviews. The superiority of gC2S over C2S increases as the lengths of reviews increase. }
	\label{fig::lm-length}
\end{figure}

\subsection{Language Modeling}

We start with the task of language modeling. Table~\ref{tab::results-lm} compares the performance of language modeling with the approach RNN, C2S and gC2S. First, both the C2S and gC2S with either the sentiment context, product context or their combination outperform the vanilla RNN without context information. This shows that contextual information are indeed helpful for natural language understanding and prediction. Second, the product context seems to be more informative than the sentiment context for language modeling, no matter with the C2S or gC2S model. This may be that there are many words in the reviews that are relevant to the product information. Comparing the C2S and gC2S model, the gC2S model consistently outperforms the C2S model no matter which contexts are used. As explained previously, this is because in the C2S model, the context information may not be able to affect the words that are far away from the beginning of the sequences while the gC2S effectively addresses this through adding the direct dependency between the contexts and the words in the sequences.  

To further verify this, we compare the results of C2S and gC2S on different lengths of reviews. Fig.~\ref{fig::lm-length} presents the results. We can see that as the lengths of the reviews increase, the gC2S model outperforms the C2S model more significantly, showing its effectiveness for modeling lengthy reviews.

Fig.~\ref{fig::gating-cs} presents several examples on the test data showing that which words are affected by the contexts. We mark the words with the largest gating values (the average of the gating vector is compared here). We can see that most of the words strongly affected by the contexts are words related to the products or sentiments. 

Overall, we can see that the gC2S model is indeed more effective than C2S. Therefore, in all the following experiments, we only use gC2S. 

\subsection{Fake Review Detection}
To further evaluate the effectiveness of the gC2S model for natural language generation, we choose the task of fake review detection, which aims to classify whether the reviews are written by real users or generated by the gC2S model. Real reviews are treated as positive, and fake reviews are treated as negative. For the evaluation data, we randomly select some products which have at least two real reviews for each rating score in the \emph{Amazon} data and one review for each rating score in the \emph{TripAdvisor} data. For each real review, a fake review is generated with gC2S according to its contexts. Table~\ref{tab::fake-test-data} summarizes the number of reviews in each domain.

\begin{table}[!htdb]
	\caption{Summary of the evaluation data for fake review detection.}
	\label{tab::fake-test-data}
	\centering
	\scalebox{0.9}{
		\begin{tabular}{c|c|c|c} \hline \hline
			Domain& \#products & \#reviews/rating & total \\ \hline
			Book&74&2&740        \\ \hline		
			Electronic&100&2&1,000        \\ \hline
			Movie&100&2&1,000    \\ \hline
			Hotel&55&1&275     \\ \hline \hline
		\end{tabular}
	}
\end{table}

\begin{table}[!htdb]
	\caption{{Results of fake review classification with human judgments (Positive: real reviews, Negative: fake reviews). In all the domains, more than 50\% of the fake reviews are misclassified. }}
	\label{tab::result-human}
	\centering
	\scalebox{0.95}{
		\begin{tabular}{c|c|c|c|c} \hline \hline
			Domain& TP & FN &TN &FP \\ \hline
			Book&73.9&26.1&47.1&52.9        \\ \hline
			Electronic&77.2&22.8&43.5&56.5    \\ \hline
			Movie &79.7&20.3&46.1&53.9       \\ \hline
			Hotel&85.2&14.8&48.5&51.5    \\ \hline
			Overall&77.9&22.1&47.5&52.5 \\ \hline \hline 
		\end{tabular}
	}
	\\ \scriptsize TP: true positive, FP: false positive, TN: true negative, FN: false negative	
\end{table}

\noindent \textbf{Human Evaluation.} We use the Amazon Mechanical Turk to evaluate whether the reviews are fake not. We divide all the data into different batches. Each batch contains twenty reviews about the same product. We show the urls of the products, the sentiment rating and the review content to users to ask the turkers to judge whether the reviews are written by real users or not. To control the quality of the results, some ``gotcha'' questions are inserted in the middle of the list of reviews. Only the results judged by users who answer the ``gotcha'' questions correctly are kept. The kappa score is .... We summarize the final results into Table~\ref{tab::result-human}.

We can see that in all the domains, more than 50\% of the fake reviews generated by the gC2S model are misclassified by the Turkers, and around 80\% of the real reviews are correctly classified. This shows that the reviews generated by the gC2S model are indeed very natural.

\begin{figure}[htbp]
	\centering
	\includegraphics[width=0.25\textwidth]{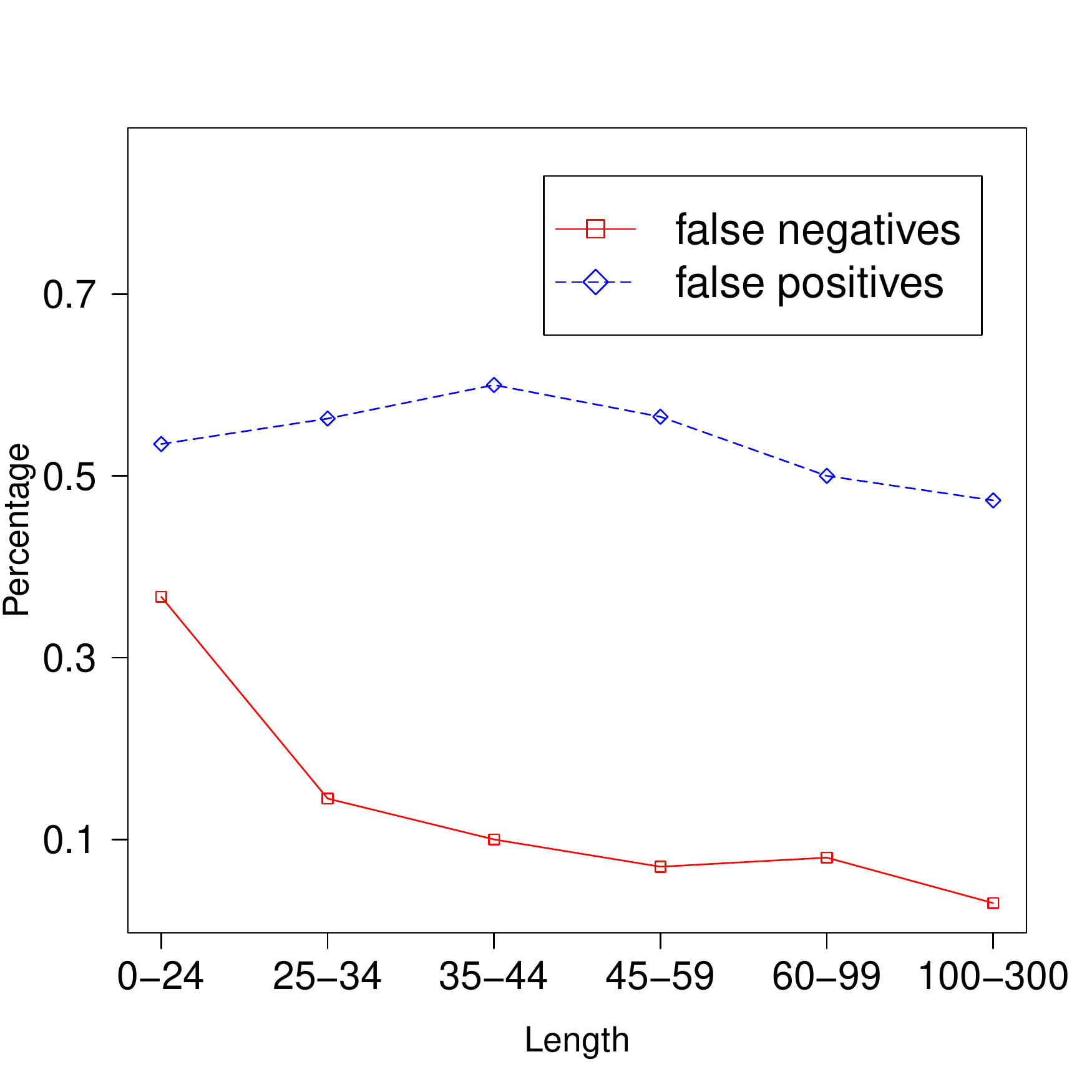}		
	\caption{ Results of human evaluation w.r.t. different lengths of reviews. Lengthy fake reviews are more likely to be detected by real users.  }
	\label{fig::hj-length}
\end{figure}


As more samples are generated by the RNN, more mistakes are likely to make by the model. Therefore, we want to see how the results of human evaluate change as the lengths of the reviews increase. Fig.~\ref{fig::hj-length} presents the human evaluation results w.r.t different lengths of reviews. We can see that for both the lengthy fake and real reviews, fewer percentages are misclassified by human judges. However, we can see that even for the fake reviews with more than 150 words, around 40\% of them are still misclassified by the human judges.

\begin{table}[!htdb]
	\caption{Results of fake review detection in \emph{TripAdvisor} with the approach in ~\cite{ott2011finding}. More than 90\% fake reviews are misclassified.}
	\label{tab::fake-classifier}
	\centering
	\scalebox{1.0}{
		\begin{tabular}{c|c|c|c|c} \hline \hline
			Feature& TP & FN &TN &FP \\ \hline
			Unigram+LR&88.8&11.2&8.3&91.7        \\ \hline
			Bigram+LR&87.3&12.7&9.1&90.9    \\ \hline \hline
		\end{tabular}
	}
	\\ \scriptsize LR: logistic regression classifier
\end{table}

\noindent \textbf{Automatic Classification.} Another way to evaluate the effectiveness of our approach is to see how well existing state-of-the-art fake review detection algorithm performs on our generated fake reviews. We adopt the approach in~\cite{ott2011finding}, which trains a classifier with 800 real reviews from \emph{TripAdvisor} and 800 fake reviews written by the Amazon Mechanical Turkers\footnote{The training data is available at~\url{http://myleott.com/op_spam/}.}. Here we use the unigram and bigram features for classification, with which the results are very close to the best results according to~\cite{ott2011finding}. Table~\cite{tab::fake-classifier} summarizes the results. We can see that more than 90\% of the fake reviews generated by the gC2S model are misclassified as the real reviews by the classifier.  

\begin{table}[!htdb]
	\caption{Fine granularity sentiment classification. Results on the true and fake reviews are very close.}
	\label{tab::sentiment-5}
	\centering
	\scalebox{0.9}{
		\begin{tabular}{c|c|c|c|c} \hline \hline
			Domain&Data&Precision&Recall&F1   \\ \hline
			\multirow{ 2}{*}{Book}& True     &  0.533  & 0.452 & 0.480\\ \cline{2-5}
			& Fake     &  0.574  & 0.501 & 0.529  \\ \hline
			\multirow{ 2}{*}{Electronic}& True     &  0.450  & 0.451 & 0.410\\ \cline{2-5}
			& Fake     &  0.450  & 0.461 & 0.419  \\ \hline									
			\multirow{ 2}{*}{Movie}& True     &  0.536  & 0.443 & 0.474\\ \cline{2-5}
			& Fake     &  0.593  & 0.472 & 0.507  \\ \hline						
			\multirow{ 2}{*}{Hotel}& True     &  0.562  & 0.504 & 0.527\\ \cline{2-5}
			& Fake     &  0.575  & 0.361 & 0.396  \\ \hline \hline														  		                      						
		\end{tabular}
	}
\end{table}

\begin{table}[!htdb]
	\caption{Binary sentiment classification. Results on the true and fake reviews are very close.}
	\label{tab::sentiment-2}	
	\centering
	\scalebox{0.9}{
		\begin{tabular}{c|c|c|c|c} \hline \hline
			Domain&Data&Precision&Recall&F1   \\ \hline
			\multirow{ 2}{*}{Book}& True     &  0.963  & 0.982 & 0.972\\ \cline{2-5}
			& Fake     & 0.971   & 0.994 &  0.982 \\ \hline
			\multirow{ 2}{*}{Electronic}& True     &  0.781  & 0.953 & 0.858\\ \cline{2-5}
			& Fake     &  0.761  & 0.993 &  0.861 \\ \hline									
			\multirow{ 2}{*}{Movie}& True     &  0.973  & 0.983 & 0.978\\ \cline{2-5}
			& Fake     &   0.973 & 0.996 &  0.984 \\ \hline						
			\multirow{ 2}{*}{Hotel}& True     &  0.985  & 0.988 & 0.987\\ \cline{2-5}
			& Fake     &  0.947  & 1.000 & 0.973  \\ \hline \hline														  		                      						
		\end{tabular}
	}
\end{table}

\subsection{Sentiment Classification}

The above results show that given particular contexts, the gC2S model is able to generate very natural reviews that are indistinguishable from real reviews. But how well the generated reviews reflect the context information, e.g., how well the generated reviews express the sentiment polarity?  Therefore, in this part we compare the results of sentiment classification on both the fake and real reviews. Two types of sentiment classification are conducted: finer granularity, i.e., sentiments with five different rating scores, and binary classification, in which reviews with 4 and 5 ratings are treated as positive and reviews with 1 and 2 ratings are treated as negative.

To conduct the classification, we randomly sample one 100,000 real reviews from the training data for training the sentiment classifier. As for the evaluation data, a fake review is generated for each test data according to its contexts. Table~\ref{tab::sentiment-5} and~\ref{tab::sentiment-2} summarize the results of fine granularity and binary sentiment classification respectively. We can see that the results on the real and fake reviews are very close to each other. On some domains (e.g., \emph{Book} and \emph{Movie}), the results on the fake reviews are even better than the real reviews, showing that the reviews generated by the gC2S model accurately reflect the sentiment polarity.

%

Finally, we present some examples of fake reviews generated by the gC2S model (Table~\ref{tab::fake-reviews}). We can see that the generated reviews are very natural, which are grammatically correct, accurately reflect the sentiment polarity and product information.

\begin{table*}[!htdb]
	\caption{Examples of fake reviews generated by the gC2S model.}
	\label{tab::fake-reviews}
	\centering
	\scalebox{0.9}{
		\begin{tabular}{c|c|c|c}  \hline \hline
		Domain & Product & Rating & Review	\\ \hline
		\multirow{2}{*}{ Movie}&\multirow{2}{*}{``Frozen''} &1& \multicolumn{1}{p{10cm}}{i love disney movies but this one was not at all what i expected . the story line was so predictable , it was dumb .}\\ \cline{3-4}
		&&3& \multicolumn{1}{p{10cm}}{i liked the movie but it didn't hold my attention as much as i expected . they just don't make movies like this anymore .}\\\cline{3-4}
		&&5& \multicolumn{1}{p{10cm}}{my son loves this movie and it is a good family movie . i would recomend it for anyone who likes to watch movies with kids .}\\\cline{3-4}		
		 \hline	
		\multirow{2}{*}{Electronic}&\multirow{2}{1in}{ ``Leather case cover for ipad mini''} &1& \multicolumn{1}{p{10cm}}{i bought this case for my ipad 3 . it was not as pictured and it was too small and the ipad mini was not secured inside it , so i returned it .}\\ \cline{3-4}
		&&3& \multicolumn{1}{p{10cm}}{the case is good for the price , but seems to be of very thin plastic and not well made . i use the stylus for reading . i would recommend it if you have a mini ipad .}\\\cline{3-4}
		&&5& \multicolumn{1}{p{10cm}}{the cover is very good and it fits the ipad mini perfectly and the color is exactly what i was looking for .}\\\cline{3-4}		
		\hline				 														  		                      				
		\end{tabular}
	}
\end{table*}

%% file: conclusion.tex
\section{Conclusion}
\label{sec::conclusion}
This paper studied context-aware natural language generation. We proposed two approaches, C2S and gC2S, which encode the contexts into semantic representations and then decode the representations into text sequences. The gC2S model significantly outperforms the C2S model as it adds skip-connections between the context representations and the words in the sequences, allowing the information from the contexts to be able to directly affect the generation of words. We evaluated our approaches on the user reviews data. Experimental results show that more than 50\% of the fake reviews generated by our approach are misclassified by human judges, and more than 90\% of the reviews are misclassified by existing fake review detection algorithm. 

In the future, we plan to integrate more context information, e.g., the user, the detailed descriptions of the products, the product prices, into our approaches and also evaluate our approaches in other scenarios, e.g., generating the titles of scientific papers based on the author, venue, and time information. It may be also beneficial to improve our model through the attention mechanical~\cite{bahdanau2014neural}, i.e., attending to different types of contexts when generating words in different positions.

%% file: review_generation_aaai17.bbl
\begin{thebibliography}{}

\bibitem[\protect\citeauthoryear{Bahdanau, Cho, and
  Bengio}{2014}]{bahdanau2014neural}
Bahdanau, D.; Cho, K.; and Bengio, Y.
\newblock 2014.
\newblock Neural machine translation by jointly learning to align and
  translate.
\newblock {\em arXiv preprint arXiv:1409.0473}.

\bibitem[\protect\citeauthoryear{Bowman \bgroup et al\mbox.\egroup
  }{2015}]{bowman2015generating}
Bowman, S.~R.; Vilnis, L.; Vinyals, O.; Dai, A.~M.; Jozefowicz, R.; and Bengio,
  S.
\newblock 2015.
\newblock Generating sentences from a continuous space.
\newblock {\em arXiv preprint arXiv:1511.06349}.

\bibitem[\protect\citeauthoryear{Cao \bgroup et al\mbox.\egroup
  }{2009}]{cao2009context}
Cao, H.; Hu, D.~H.; Shen, D.; Jiang, D.; Sun, J.-T.; Chen, E.; and Yang, Q.
\newblock 2009.
\newblock Context-aware query classification.
\newblock In {\em Proceedings of the 32nd international ACM SIGIR conference on
  Research and development in information retrieval},  3--10.
\newblock ACM.

\bibitem[\protect\citeauthoryear{Cheyer and Guzzoni}{2014}]{cheyer2014method}
Cheyer, A., and Guzzoni, D.
\newblock 2014.
\newblock Method and apparatus for building an intelligent automated assistant.
\newblock US Patent 8,677,377.

\bibitem[\protect\citeauthoryear{Cho \bgroup et al\mbox.\egroup
  }{2014}]{cho2014learning}
Cho, K.; Van~Merri{\"e}nboer, B.; Gulcehre, C.; Bahdanau, D.; Bougares, F.;
  Schwenk, H.; and Bengio, Y.
\newblock 2014.
\newblock Learning phrase representations using rnn encoder-decoder for
  statistical machine translation.
\newblock {\em arXiv preprint arXiv:1406.1078}.

\bibitem[\protect\citeauthoryear{Graves}{2013}]{graves2013generating}
Graves, A.
\newblock 2013.
\newblock Generating sequences with recurrent neural networks.
\newblock {\em arXiv preprint arXiv:1308.0850}.

\bibitem[\protect\citeauthoryear{Hochreiter and
  Schmidhuber}{1997}]{hochreiter1997long}
Hochreiter, S., and Schmidhuber, J.
\newblock 1997.
\newblock Long short-term memory.
\newblock {\em Neural computation} 9(8):1735--1780.

\bibitem[\protect\citeauthoryear{Karpathy, Johnson, and
  Li}{2015}]{karpathy2015visualizing}
Karpathy, A.; Johnson, J.; and Li, F.-F.
\newblock 2015.
\newblock Visualizing and understanding recurrent networks.
\newblock {\em arXiv preprint arXiv:1506.02078}.

\bibitem[\protect\citeauthoryear{Mei \bgroup et al\mbox.\egroup
  }{2006}]{mei2006probabilistic}
Mei, Q.; Liu, C.; Su, H.; and Zhai, C.
\newblock 2006.
\newblock A probabilistic approach to spatiotemporal theme pattern mining on
  weblogs.
\newblock In {\em Proceedings of the 15th international conference on World
  Wide Web},  533--542.
\newblock ACM.

\bibitem[\protect\citeauthoryear{Mikolov and Zweig}{2012}]{mikolov2012context}
Mikolov, T., and Zweig, G.
\newblock 2012.
\newblock Context dependent recurrent neural network language model.
\newblock In {\em SLT},  234--239.

\bibitem[\protect\citeauthoryear{Oh and Rudnicky}{2000}]{oh2000stochastic}
Oh, A.~H., and Rudnicky, A.~I.
\newblock 2000.
\newblock Stochastic language generation for spoken dialogue systems.
\newblock In {\em Proceedings of the 2000 ANLP/NAACL Workshop on Conversational
  systems-Volume 3},  27--32.
\newblock Association for Computational Linguistics.

\bibitem[\protect\citeauthoryear{Ott \bgroup et al\mbox.\egroup
  }{2011}]{ott2011finding}
Ott, M.; Choi, Y.; Cardie, C.; and Hancock, J.~T.
\newblock 2011.
\newblock Finding deceptive opinion spam by any stretch of the imagination.
\newblock In {\em Proceedings of the 49th Annual Meeting of the Association for
  Computational Linguistics: Human Language Technologies-Volume 1},  309--319.
\newblock Association for Computational Linguistics.

\bibitem[\protect\citeauthoryear{Rush, Chopra, and
  Weston}{2015}]{rush2015neural}
Rush, A.~M.; Chopra, S.; and Weston, J.
\newblock 2015.
\newblock A neural attention model for abstractive sentence summarization.
\newblock {\em arXiv preprint arXiv:1509.00685}.

\bibitem[\protect\citeauthoryear{Sordoni \bgroup et al\mbox.\egroup
  }{2015}]{sordoni2015neural}
Sordoni, A.; Galley, M.; Auli, M.; Brockett, C.; Ji, Y.; Mitchell, M.; Nie,
  J.-Y.; Gao, J.; and Dolan, B.
\newblock 2015.
\newblock A neural network approach to context-sensitive generation of
  conversational responses.
\newblock {\em arXiv preprint arXiv:1506.06714}.

\bibitem[\protect\citeauthoryear{Sutskever, Martens, and
  Hinton}{2011}]{sutskever2011generating}
Sutskever, I.; Martens, J.; and Hinton, G.~E.
\newblock 2011.
\newblock Generating text with recurrent neural networks.
\newblock In {\em Proceedings of the 28th International Conference on Machine
  Learning (ICML-11)},  1017--1024.

\bibitem[\protect\citeauthoryear{Wang and Cho}{2015}]{wang2015larger}
Wang, T., and Cho, K.
\newblock 2015.
\newblock Larger-context language modelling.
\newblock {\em arXiv preprint arXiv:1511.03729}.

\bibitem[\protect\citeauthoryear{Wen \bgroup et al\mbox.\egroup
  }{2015a}]{wen2015stochastic}
Wen, T.-H.; Gasic, M.; Kim, D.; Mrksic, N.; Su, P.-H.; Vandyke, D.; and Young,
  S.
\newblock 2015a.
\newblock Stochastic language generation in dialogue using recurrent neural
  networks with convolutional sentence reranking.
\newblock {\em arXiv preprint arXiv:1508.01755}.

\bibitem[\protect\citeauthoryear{Wen \bgroup et al\mbox.\egroup
  }{2015b}]{wen2015semantically}
Wen, T.-H.; Gasic, M.; Mrksic, N.; Su, P.-H.; Vandyke, D.; and Young, S.
\newblock 2015b.
\newblock Semantically conditioned lstm-based natural language generation for
  spoken dialogue systems.
\newblock {\em arXiv preprint arXiv:1508.01745}.

\bibitem[\protect\citeauthoryear{Xu \bgroup et al\mbox.\egroup
  }{2015}]{xu2015show}
Xu, K.; Ba, J.; Kiros, R.; Courville, A.; Salakhutdinov, R.; Zemel, R.; and
  Bengio, Y.
\newblock 2015.
\newblock Show, attend and tell: Neural image caption generation with visual
  attention.
\newblock {\em arXiv preprint arXiv:1502.03044}.

\bibitem[\protect\citeauthoryear{Zhang and Lapata}{2014}]{zhang2014chinese}
Zhang, X., and Lapata, M.
\newblock 2014.
\newblock Chinese poetry generation with recurrent neural networks.
\newblock In {\em EMNLP},  670--680.

\end{thebibliography}
